%% file: main.tex
\documentclass[9pt]{article}
\pdfoutput=1
\usepackage{spconf}
\usepackage[english]{babel}%
\usepackage[utf8]{inputenc} %
\usepackage{csquotes}%
\usepackage[backend=biber, 
            style=numeric,
            isbn=false,
            url=true,
            doi=false]{biblatex}
\AtEveryBibitem{%
  \clearfield{note}%
}

\usepackage{xspace} %
\usepackage{bm,bbm}
\usepackage{mathtools} %
\usepackage{amssymb}
\DeclareMathAlphabet{\mathcal}{OMS}{cmsy}{m}{n}
\usepackage{subfig}

\usepackage[style=base,labelfont=bf]{caption}
\usepackage{booktabs} %
\usepackage{tabularx} %
\usepackage{multirow}
\ifpdf
   \usepackage{graphicx}
\else
   \usepackage{graphicx}
\fi
\usepackage[T1]{fontenc}    %
\usepackage{xcolor}
\usepackage{hyperref}       %
\usepackage{placeins}
\usepackage{url}            %
\usepackage{booktabs}       %
\usepackage{amsfonts}       %
\usepackage{nicefrac}       %
\usepackage{microtype}      %
\usepackage{tikz}
\usetikzlibrary{dsp,chains}

\usepackage[inline]{enumitem}
\usepackage{etoolbox}
\robustify{\subref}
\usepackage{breqn}
\input{defs}

\title{Deep learning in the wavelet domain}

\name{Fergal Cotter and Prof Nick Kingsbury}
\address{Signal Processing, Engineering Department\\
         University of Cambridge, U.K.\\
         \texttt{fbc23@cam.ac.uk}, \texttt{ngk@eng.cam.ac.uk}}

\begin{document}

\maketitle

\input{abstract.tex}
\input{intro.tex}
\input{section2.tex}
\input{section3.tex}

\section*{References}
\printbibliography[heading=none]
\end{document}

%% file: defs.tex
\newcommand{\CWT}{\ensuremath{{\mathbb{C}}\mathrm{WT}}\xspace} %
\newcommand{\DTCWT}{{\ensuremath{\mathrm{DT}\CWT}}\xspace}

\newcommand{\x}{\times}                     %
\newcommand{\degs}{{\ensuremath{^{\circ}}}\xspace}

\newcommand{\F}[1]{\ensuremath{\mathrm{#1}}\xspace}

\newcommand{\dydx}[2]{\frac{\partial #1}{\partial #2}}

\newcommand{\reals}[1][]{\ensuremath{{\mathbb{R}}^{#1}}\xspace}
\newcommand{\complexes}[1][]{\ensuremath{{\mathbb{C}}^{#1}}\xspace}

\usepackage{catoptions}
\makeatletter

\def\Autoref#1{%
  \begingroup
  \edef\reserved@a{\cpttrimspaces{#1}}%
  \ifcsndefTF{r@#1}{%
    \xaftercsname{\expandafter\testreftype\@fourthoffive}
      {r@\reserved@a}.\\{#1}%
  }{%
    \ref{#1}%
  }%
  \endgroup
}
\def\testreftype#1.#2\\#3{%
  \ifcsndefTF{#1autorefname}{%
    \def\reserved@a##1##2\@nil{%
      \uppercase{\def\ref@name{##1}}%
      \csn@edef{#1autorefname}{\ref@name##2}%
      \autoref{#3}%
    }%
    \reserved@a#1\@nil
  }{%
    \autoref{#3}%
  }%
}
\makeatother

%% file: abstract.tex
\begin{abstract}
This paper examines the possibility of, and the possible advantages to learning the
filters of convolutional neural networks (CNNs) for image analysis in the wavelet domain.  
We are stimulated by both Mallat's scattering transform \cite{mallat_group_2012} and the idea of
filtering in the Fourier domain. It is important to explore new spaces in which to learn, as these
may provide inherent advantages that are not available in the pixel space. However, the scattering
transform is limited by its inability to learn in between scattering orders, and any Fourier domain
filtering is limited by the large number of filter parameters needed to get localized filters.
Instead we consider filtering in the wavelet domain with learnable filters. The wavelet space allows
us to have local, smooth filters with far fewer parameters, and learnability can give us
flexibility. 

We present a novel layer which takes CNN activations into the wavelet space, learns
parameters and returns to the pixel space. This allows it to be easily dropped in to any neural
network without affecting the structure. As part of this work, we show how to pass gradients through 
a multirate system and give preliminary results.

\end{abstract}

%% file: intro.tex
\section{Introduction}\label{sec:intro}
Using wavelet based methods with deep learning is nascent but not novel. Wavelets have been applied 
to texture classification \cite{fujieda_wavelet_2017, sifre_combined_2012}, super-resolution
\cite{guo_deep_2017} and for adding detail back into dense pixel-wise segmentation tasks
\cite{ma_detailed_2018}. One exciting piece of work built on wavelets is the Scattering Transform
\cite{mallat_group_2012}, which has been used as a feature extractor for learning, firstly with
simple classifiers \cite{bruna_invariant_2013, singh_scatternet_2017}, and later as a front end to
hybrid deep learning tasks\cite{oyallon_scaling_2017, singh_scatternet_2018}. Despite their power
and simplicity, scattering features are fixed and are visibly different to regular CNN features
\cite{cotter_visualizing_2017} - their nice invariance properties come at the cost of flexibility,
as there is no ability to learn in between scattering layers. 

For this reason, we have been investigating a slightly different approach, more similar to the Fourier based work in
\cite{rippel_spectral_2015} in which Rippel et.\ al.\ investigate parameterization of filters in the Fourier
domain. In the forward pass, they take the inverse DFT of their filter,
and then apply normal pixel-wise convolution. We wish to extend this by not only parameterizing
filters in the wavelet domain, but by performing the convolution there as well
(i.e., also taking the activations into the wavelet domain). After 
processing is done, we can return to the pixel domain. Doing these forward
and inverse transforms has two significant advantages: 
\begin{enumerate*}[label=\roman*)]
  \item the layers can easily replace standard convolutional layers if they accept and return the
    same format;
  \item we can learn both in the wavelet and pixel space.
\end{enumerate*}

As neural network training involves presenting thousands of training samples, we want our layer to
be fast. To achieve this we would ideally choose to use a critically sampled filter bank
implementation. The fast 2-D Discrete Wavelet Transform (DWT) is a possible option, but it has two
drawbacks: it has poor directional selectivity and any alteration of wavelet coefficients will cause
the aliasing cancelling properties of the reconstructed signal to disappear.  Instead we choose to
use the Dual-Tree Complex Wavelet Transform (\DTCWT) \cite{selesnick_dual-tree_2005} as at the
expense of limited redundancy (4:1), it enables us to have better directional selectivity, and allows
us to modify the wavelet coefficients and still have minimal aliasing terms when we reconstruct
\cite{kingsbury_complex_2001}.

\Autoref{sec:method} of the paper describes the implementation details of our design, and
\autoref{sec:results} describes the experiments and results we have done so far.

%% file: section2.tex
\section{Method}\label{sec:method}
\begin{figure*}[ht]
  \centering
  \subfloat[]{%
    \includegraphics[height=6cm]{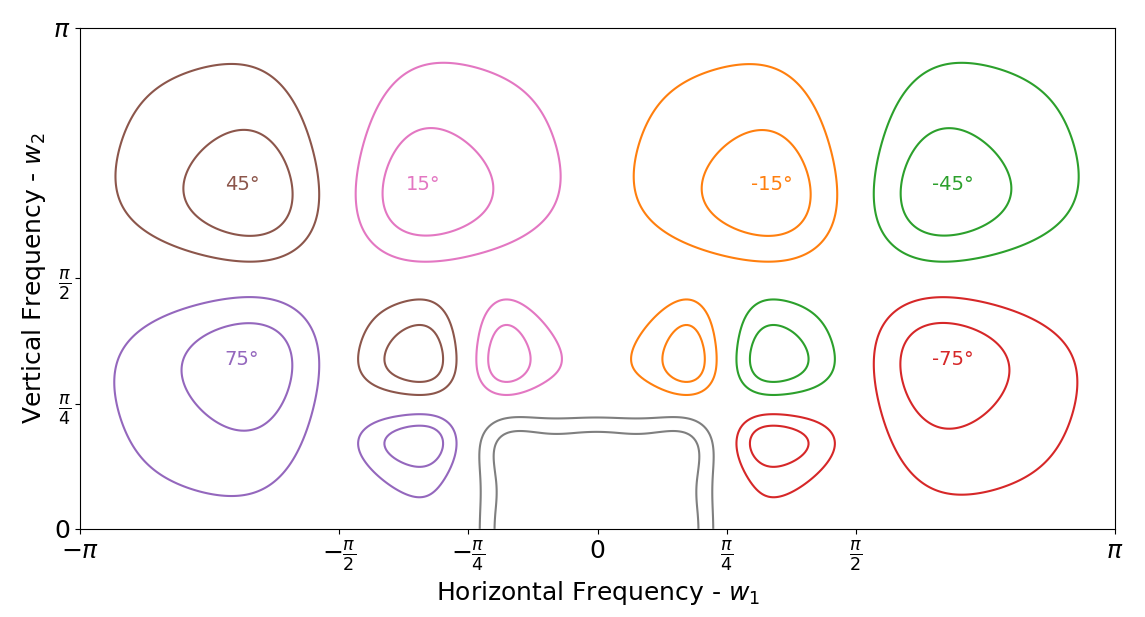}
    \label{fig:dtcwt_bands_freq}
  }
  \hspace{1cm}
  \subfloat[]{%
    \includegraphics[height=5.7cm]{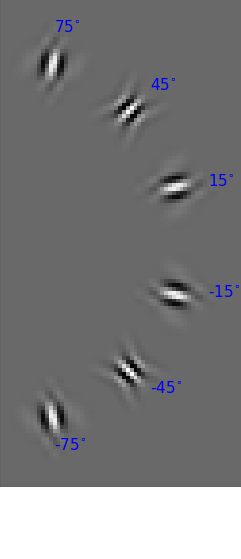}
    \label{fig:dtcwt_bands_impulse}
  }
  \newline
  \subfloat[]{%
    \includegraphics[width=.8\textwidth]{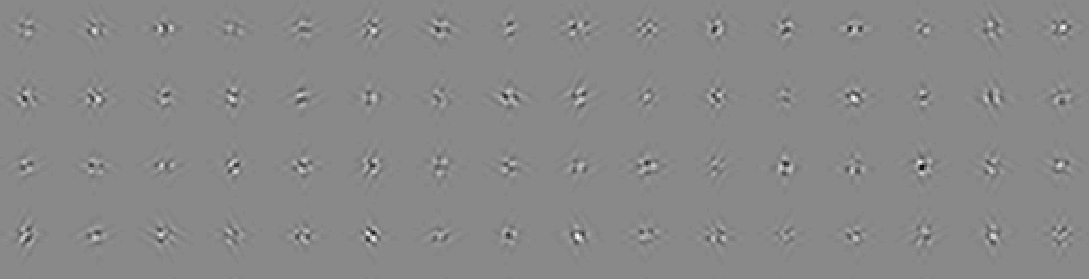}
    \label{fig:example_impulses}
  }
  \caption{\subref{fig:dtcwt_bands_freq} Contour plots at -1dB and -3dB showing the support in the
    Fourier domain of the 6 subbands of the \DTCWT at scales 1 and 2 and the scale 2 lowpass. These
    are the product $P(z)Q(z)$ from \autoref{eq:end_to_end1}.%
    \subref{fig:dtcwt_bands_impulse} The pixel domain impulse responses for the second scale
    wavelets. \subref{fig:example_impulses} Example impulses of our layer when $g_1$, and $g_{lp}$ are 0 
    and $g_2 \in \mathbb{C}^{6\x 1\x 1}$, with each real and imaginary element drawn 
    from $\mathcal{N}(0,1)$. I.e., only information in the 6 subbands with $\frac{\pi}{4} < |w_1|,
    |w_2| < \frac{\pi}{2}$ from \subref{fig:dtcwt_bands_freq} is passed through.} 
  \label{fig:dtcwt_bands}
  \vspace{-5mm}
\end{figure*}
In a standard convolutional layer, an input with $C$ channels, $H$ rows and $W$ columns is $X \in
\reals[C\x H\x W]$, which is then convolved with $F$ filters of spatial size $K$ - $w \in \reals[F
\x C\x K\x K]$, giving $Y \in \reals[F\x H\x W]$. In many systems like
\cite{krizhevsky_imagenet_2012, he_deep_2015}, the first layer is typically a selection of bandpass
filters, selecting edges with different orientations and center frequencies. In the wavelet space
this would be trivial - take a decomposition of each input channel and keep individual subbands (or
equivalently, attenuate other bands), then take the inverse wavelet transform.
\autoref{fig:dtcwt_bands} shows the frequency space for the \DTCWT and makes it clearer as to how
this could be done practically for a two scale transform. To attenuate all but say the $15 \degs$
band at the first scale for the first input channel, we would need to have $13C$ gains for the 13
subbands and $C$ input channels, $13C-1$ of which would be zero and the remaining coefficient one.

Instead of explicitly setting the gains, we can randomly initialize them and use backpropagation to
learn what they should be. This gives us the power to learn more complex shapes rather than simple
edges, as we can mix the regions of the frequency space per input channel in an arbitrary way. 

\subsection{Memory Cost}
Again considering a two scale transform --- instead of learning $w\in \reals[F \x C\x K\x K]$ we
learn complex gains at the two scales, and a real gain for the real lowpass:
$$\left\{g_1 \in \complexes[F\x C\x 6\x 1\x 1], g_2 \in \complexes[F\x C\x 6\x 1\x 1], g_{lp} \in \reals[F\x C\x 1\x 1]
  \right\}$$ 
We have set the spatial dimension to be $1\x 1$ to show that this gain is identical to a $1\x 1$ 
convolution over the complex wavelet coefficients. If we wish, we can learn larger spatial sizes to have more complex
attenuation/magnification of the subbands. We also can use more/fewer than 2 wavelet scales.  At first
glance, we have increased our parameterization by a factor of 25 (13 subbands, of which all but the
lowpass are complex), but each one of these gains affects a large spatial size. For the first scale,
the effective size is about $5\x 5$ pixels, for the second scale it is about $15\x 15$.

\begin{figure*}[ht]
  \centering
  \subfloat[]{%
    \includegraphics[width=.85\textwidth]{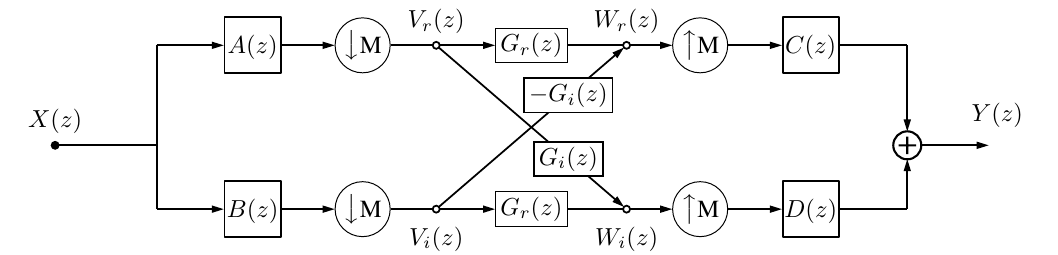}
    \label{fig:fwd_pass}
  }
   \newline
  \subfloat[]{%
    \includegraphics[width=.85\textwidth]{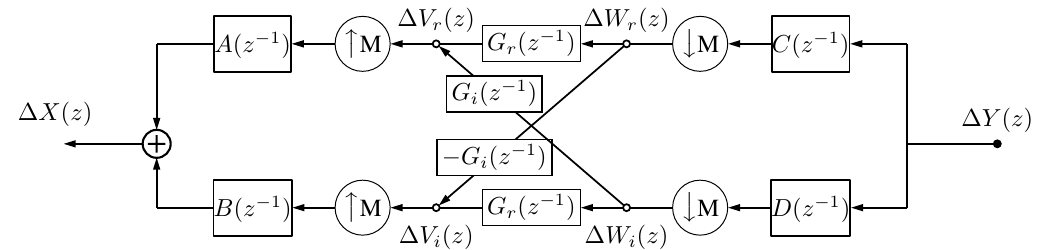}
    \label{fig:bwd_pass}
  }
  \caption{\subref{fig:fwd_pass} Forward and \subref{fig:bwd_pass} backward pass of our system, based on Figure~4 in
  \cite{kingsbury_complex_2001}. Ignoring the $G$ gains, the top and bottom paths (through $A, C$
  and $B, D$ respectively) make up the the real and imaginary parts for \emph{one subband} of the dual tree system.
  Combined, $A+jB$ and $C-jD$ make the complex filters necessary to have support on one side of the
  Fourier domain (see \autoref{fig:dtcwt_bands}). Adding in the complex gain $G_r + jG_i$, we can now
  attenuate/shape the impulse response in each of the subbands. To allow for learning, we need
  backpropagation. The bottom diagram indicates how to pass gradients $\Delta Y(z)$ through the
  layer. Note that upsampling has become downsampling, and convolution has become convolution with
  the time reverse of the filter (represented by $z^{-1}$ terms).}
  \label{fig:fwd_bwd}
  \vspace{-5mm}
\end{figure*}
\subsection{Computational Cost}
A standard convolutional layer needs $K^2 F$ multiplies per input pixel (of which there are $C\x H\x
W$). In comparison, the wavelet gain method does a set number of operations per pixel for the
forward and inverse transforms, and then applies gains on subsampled activations. For a 2 level
\DTCWT the transform overhead is about 60 multiplies for both the forward and inverse transform. It
is important to note that unlike the filtering operation, this does not scale with $F$. The learned
gains in each subband do scale with the number of output channels, but can have smaller spatial size
(as they have larger effective sizes) as well as having fewer pixels to operate on (because of the
decimation). The end result is that as $F$ and $C$ grow, the overhead of the $C$ forward and $F$
inverse transforms is outweighed by cost of $FC$ mixing processes, which should in turn be
significantly less than the cost of $FC$ $K\x K$ standard convolutions for equivalent spatial sizes.
\subsection{Examples}
\autoref{fig:example_impulses} show example impulse responses of our layer. These impulses were generated by
randomly initializing both the real and imaginary parts of $g_2 \in \complexes[6\x 1\x 1]$ 
from $\mathcal{N}(0,1)$ and $g_1, g_{lp}$ are set to 0. I.e. each shape has 12 random variables. It
is good to see that there is still a large degree of variability between shapes. Our experiments
have shown that the distribution of the normalized cross-correlation between 512 of such randomly generated shapes
matches the distribution for random vectors with roughly 11.5 degrees of freedom.
\subsection{Forward propagation}
\autoref{fig:fwd_bwd} shows the block diagram using $Z$-transforms for a single band of our system (it
is based on Figure~4 in \cite{kingsbury_complex_2001}). To keep things simple for the rest of
\autoref{sec:method} the figure shown is for a 1-D system; 
it is relatively straightforward to extend this to 2-D\cite{selesnick_dual-tree_2005}. The complex
analysis filter (taking us into the wavelet domain) is $P(z) = \frac{1}{2}\left(A(z)+jB(z)\right)$ 
and the complex synthesis filter (returning us to the pixel domain) is 
$Q(z) = \frac{1}{2}\left(C(z) - jD(z)\right)$ 
where $A,B,C,D$ are real.  If $G(z) = G_r(z) + jG_i(z) = 1$ then the end-to-end transfer function is (from section 4 of
\cite{kingsbury_complex_2001}):
\begin{equation}\label{eq:end_to_end1}
\frac{Y(z)}{X(z)} = \frac{2}{M}\left(P(z)Q(z) + P^*(z)Q^*(z)\right)
\end{equation}
where $P, Q$ have support only in the top half of the Fourier plane and $P^*, Q^*$ are $P$ and $Q$ 
reflected in the horizontal frequency axis. Examples of $P(z)Q(z)$ for different subbands of a 2-D
\DTCWT have spectra shown in \autoref{fig:dtcwt_bands_freq}, $P^*(z)Q^*(z)$ make up the missing half
of the frequency space.\\
Modifying this from the standard wavelet equations by adding the subband gains $G_r(z)$ and
$G_i(z)$, the transfer function becomes:
\begin{equation}\label{eq:end_to_end2}
  \begin{split}
    \frac{Y(z)}{X(z)} = \frac{2}{M} \left[ \right. & G_r(z^M) \left( P(z)Q(z) + P^*(z)Q^*(z) \right) +  \\ 
     & \left. jG_i(z^M) \left(P(z)Q(z)-P^*(z)Q^*(z) \right) \right]
  \end{split}
\end{equation}

\begin{table*}[]
  \centering
{\renewcommand{\arraystretch}{1.2}
  \captionsetup{width=\textwidth}
  \caption{Comparison of LeNet with standard convolution to our proposed method which
  learns in the wavelet space (WaveLenet) on CIFAR-10 and CIFAR-100. Values reported are the average
top-1 accuracy (\%) rates for different train set sizes over 5 runs.}
\vspace{-2mm}
\begin{tabular}{cccccccc}
  \specialrule{.1em}{.1em}{.1em} 
  & Train set size & 1000 & 2000 & 5000 & 10000 & 20000 & 50000 \\ \specialrule{.1em}{.1em}{.1em} 
  \multicolumn{1}{l}{\multirow{2}{*}{CIFAR-10}} & LeNet & 
    48.5 & 52.4 & 59.5 & 65.0 & 69.5 & 73.3\\ \cline{2-8}
  \multicolumn{1}{l}{} & WaveLeNet & 
    47.3 & 52.1 & 58.7 & 63.8 & 68.0 & 72.4\\ \hline
  \multicolumn{1}{l}{\multirow{2}{*}{CIFAR-100}} & LeNet & 
    11.1 & 15.8 & 23.1 & 29.5 & 34.4 & 41.1  \\ \cline{2-8}
  \multicolumn{1}{l}{} & WaveLeNet & 
    11.1 & 15.4 & 23.2 & 28.4  & 33.9 & 39.6 \\ \specialrule{.1em}{.1em}{.1em} 
\end{tabular}\label{tab:results}
}
\end{table*}
\subsection{Backpropagation}
We start with the commonly known property that for a convolutional block, the gradient with respect to the input is the gradient with
respect to the output convolved with the time reverse of the filter. More formally, if $Y(z) = H(z) X(z)$:
\begin{equation}\label{eq:backprop}
\Delta X(z) = H(z^{-1}) \Delta Y(z)
\end{equation}
where $H(z^{-1})$ is the $Z$-transform of the time/space reverse of $H(z)$, $\Delta Y(z)
\triangleq \dydx{L}{Y}(z)$ is the gradient of the loss with respect to the output, and
$\Delta X(z) \triangleq \dydx{L}{X}(z)$ is the gradient of the loss with respect to the input. If 
H were complex, the first term in \autoref{eq:backprop} would be $\bar{H}(1/\bar{z})$, but as each individual block
in the \DTCWT is purely real, we can use the simpler form. 

Assume we already have access to the quantity $\Delta Y(z)$ (this is the input
to the backwards pass). \autoref{fig:bwd_pass} illustrates the backpropagation procedure. An
interesting result is that the backwards pass of an inverse wavelet transform is equivalent to doing
a forward wavelet transform.\footnote{As shown in \autoref{fig:bwd_pass}, the analysis and synthesis
  filters have to be swapped and time reversed. For orthogonal wavelet transforms, the synthesis
  filters are already the time reverse of the analysis filters, so no change has to be done. The q-shift filters of the
\DTCWT \cite{kingsbury_design_2003} have this property.} Similarly, the backwards pass of the forward transform is equivalent to
doing the inverse transform. The weight update gradients are then calculated by finding $\Delta W(z)
= \DTCWT\left\{ \Delta Y(z) \right\}$ and then convolving with the time reverse of the saved wavelet coefficients from
the forward pass - $V(z)$.
\begin{gather}
  \Delta G_r(z) = \Delta W_r(z) V_r(z^{-1}) + \Delta W_i(z) V_i(z^{-1})  \label{eq:gr_update}\\
  \Delta G_i(z) =  -\Delta W_r(z) V_i(z^{-1}) + \Delta W_i(z) V_r(z^{-1})  \label{eq:gi_update} 
\end{gather}
Unsurprisingly, the passthrough gradients have similar form to \autoref{eq:end_to_end2}:
\begin{equation}\label{eq:passthrough}
  \begin{split}
    \Delta X(z) = \frac{2\Delta Y(z)}{M} & \left[G_r(z^{-M})\left( PQ + P^*Q^* \right)\right. + \\
      & \left. jG_i(z^{-M}) \left(PQ-P^*Q^* \right) \right] 
\end{split}
\end{equation}
where we have dropped the $z$ terms on $P(z), Q(z), P^*(z), Q^*(z)$ for brevity.

Note that we only need to evaluate equations
\ref{eq:gr_update},\ref{eq:gi_update},\ref{eq:passthrough} over the support of $G(z)$ 
i.e., if it is a single number we only need to calculate $\left.\Delta G(z)\right\rvert_{z=0}$.

%% file: section3.tex
\section{Experiments and Preliminary Results}\label{sec:results}
To examine the effectiveness of our convolutional layer, we do a simple experiment on CIFAR-10 and
CIFAR-100. For simplicity, we compare the performance using a simple yet relatively effective convolutional
architecture - LeNet \cite{lecun_gradient-based_1998}. LeNet has 2 convolutional layers of spatial
size $5\x 5$ followed by
2 fully connected layers and a softmax final layer. We swap both these convolutional
layers out for two of our proposed wavelet gain layers (keeping the ReLU between them). As CIFAR has
very small spatial size, we only take a single scale \DTCWT\@. Therefore each gain layer has $6$ complex gains
for the 6 subbands, and a $3\x 3$ real gain for the lowpass (a total of $21C$ parameters vs $25C$
for the original system). We train both networks for 200 epochs with Adam
\cite{kingma_adam:_2014} optimizer with a constant learning rate of $10^{-3}$ and a weight decay of
$10^{-5}$. The code is available at \cite{cotter_dtcwt_2018}. \autoref{tab:results} shows the
mean of the validation set accuracies for 5 runs. The different columns represent undersampled
training set sizes (with 50000 being the full training set). When undersampling, we keep the samples
per class constant. We see our system perform only very slightly worse than the standard
convolutional layer. 

\section{Conclusion and Future Work}
In this work we have presented the novel idea of learning filters by taking activations into the
wavelet domain, learning mixing coefficients and then returning to the pixel space. This work is
done as a preliminary step; we ultimately hope that learning in both the wavelet and pixel space
will have many advantages, but as yet it has not been explored. We have considered the possible
challenges this proposes and described how a multirate system can learn through backpropagation.  

Our experiments so far have been promising. We have shown that our layer can learn in an
end-to-end system, achieving very near similar accuracies on CIFAR-10 and CIFAR-100 to the same system with
convolutional layers instead. This is a good start and shows the plausibility of such an idea, 
but we need to search for how to improve these layers if they are to be useful. 
It will be interesting to see how well we can learn on datasets with larger images
- our proposed method naturally learns large kernels, so should scale well with the image size.

In our experiments so far, we only briefly go into the wavelet domain before coming back to the
pixel domain to do ReLU nonlinearities, however we plan to explore using nonlinearities in the
wavelet domain, such as soft-shrinkage to denoise/sparsify the coefficients
\cite{donoho_ideal_1994}. We feel there are strong links between ReLU non-linearities and
denoising/sparsity ideas, and that there may well be useful performance gains from mixing real
pixel-domain non-linearities with complex wavelet-domain shrinkage functions. Thus we present these
ideas here as a starting point for a novel and exciting avenue of deep network research.